\documentclass[sigconf]{acmart}

\usepackage{balance}

\AtBeginDocument{%
  }

\copyrightyear{2026}
\acmYear{2026}
\setcopyright{cc}
\setcctype{by}
\acmConference[SIGIR '26]{Proceedings of the 49th International ACM SIGIR Conference on Research and Development in Information Retrieval}{July 20--24, 2026}{Melbourne, VIC, Australia}
\acmBooktitle{Proceedings of the 49th International ACM SIGIR Conference on Research and Development in Information Retrieval (SIGIR '26), July 20--24, 2026, Melbourne, VIC, Australia}
\acmISBN{979-8-4007-2599-9/2026/07}
\acmDOI{10.1145/3805712.3809935}
\begin{document}

\title{GATHER: Convergence-Centric Hyper-Entity Retrieval for Zero-Shot Cell-Type Annotation}

\author{Zhonghui Zhang}
\orcid{0009-0007-6935-122X}
\affiliation{%
  \institution{Artificial Intelligence Research Institute, Shenzhen University of Advanced Technology}
  \city{Shenzhen}
  \country{China}
}
\email{zhangzhonghui@suat-sz.edu.cn}

\author{Feng Jiang}
\authornote{Corresponding authors.}
\orcid{0000-0002-3465-311X}
\affiliation{%
  \institution{Artificial Intelligence Research Institute, Shenzhen University of Advanced Technology}
  \city{Shenzhen}
  \country{China}
}
\email{jiangfeng@suat-sz.edu.cn}

\author{Shaowei Qin}
\orcid{0000-0002-9774-0851}
\affiliation{%
  \institution{Artificial Intelligence Research Institute, Shenzhen University of Advanced Technology}
  \city{Shenzhen}
  \country{China}
}
\email{qinshaowei@suat-sz.edu.cn}

\author{Jiahao Zhao}
\orcid{0009-0004-8573-0826}
\affiliation{%
  \institution{Software College, Northeastern University}
  \city{Shenyang}
  \country{China}
}
\email{zhaojh3417@gmail.com}

\author{Min Yang}
\authornotemark[1]
\orcid{0000-0003-3814-2728}
\affiliation{%
  \institution{Shenzhen Institutes of Advanced Technology, Chinese Academy of Sciences; \\ Artificial Intelligence Research Institute, Shenzhen University of Advanced Technology}
  \city{Shenzhen}
  \country{China}
}
\email{min.yang@siat.ac.cn}

\renewcommand{\shortauthors}{Zhang et al.}

\begin{abstract}
Zero-shot single-cell cell-type annotation aims to determine a cell's type from a given set of expressed genes without any training.
Existing knowledge-graph-based RAG approaches retrieve evidence by expanding from source entities and relying on iterative LLM reasoning.
However, in this setting each query contains tens to hundreds of genes, where no single gene is decisive and the label emerges only from their collective co-occurrence.
Such hyper-entity queries fundamentally challenge local, entity-wise exploration strategies, which reason from individual genes, leading to poor scalability and substantial LLM cost.
We propose \textbf{GATHER} (Graph-Aware Traversal with Hyper-Entity Retrieval), a convergence-centric retriever tailored to hyper-entity queries. It performs global multi-source graph traversal and identifies topological convergence points---nodes jointly reachable from many input genes.
These convergence nodes act as high-information hyper-entities that capture entity synergy.
By incorporating node- and path-importance scoring, GATHER selects informative evidence entirely without LLM involvement during retrieval.
Instantiated on a self-constructed cell-centric biological knowledge graph (VCKG), GATHER outperforms strong KG-RAG baselines (ToG, ToG-2, RoG, PoG) on two datasets (Immune and Lung), achieving the highest exact-match accuracy (27.45\% and 59.64\%) with only a single LLM call per sample, compared to 2--61 calls for KG-RAG baselines.
Our results demonstrate that convergence nodes compress multi-entity signals into compact, high-information evidence that conveys more per item than multi-hop paths, providing an efficient global alternative to local entity-wise reasoning.

\end{abstract}



\maketitle

\section{Introduction}

Single-cell cell-type annotation~\cite{zhang2019cellmarker,aran2019reference,pasquini2021automated} aims to assign a cell type based on a cell's gene expression profile and is fundamental to computational biology, enabling downstream analyses such as cell-type discovery and disease mechanism study.
Given a cell's gene expression profile, the task relies on the \emph{joint expression pattern} of many genes rather than any single marker.
Thus, the prediction signal emerges from the \emph{global interaction} among tens to hundreds of genes.

Supervised foundation models such as scGPT~\cite{cui2024scgpt}, scBERT~\cite{yang2022scbert}, and Geneformer~\cite{theodoris2023transfer} achieve strong accuracy but operate as black boxes, limiting interpretability.
In training-free settings, large language models (LLMs)~\cite{brown2020language,achiam2023gpt} offer explainable reasoning but suffer from imprecise domain knowledge~\cite{hou2024assessing,zhao2024langcell}.
Retrieval-Augmented Generation (RAG)~\cite{lewis2020retrieval,fan2024survey} has therefore emerged as a promising paradigm to ground LLMs with structured knowledge.

However, applying RAG to cell-type annotation introduces a key challenge:
each query consists of tens to hundreds of genes, and effectively leveraging external knowledge graphs over such a large set of entities becomes non-trivial.
The central question is how to integrate structured knowledge while jointly considering all source entities. Existing knowledge graph-based RAG methods~\cite{pan2024unifying,zhu2024llms}, such as ToG~\cite{sun2024tog} and PoG~\cite{tan2025paths}, primarily adopt a \emph{local expansion} paradigm.
They start from each source entity independently, explore neighboring nodes, and aggregate retrieved paths as separate evidence.
While effective for few-entity queries, this strategy becomes problematic in hyper-entity settings.
First, independent expansion fragments the collective signal and fails to explicitly model interactions among many entities.
Second, the search and LLM-interaction costs scale rapidly with the number of source entities, expansion breadth, and depth.

\begin{figure}[t]
  \centering
  \includegraphics[width=\columnwidth]{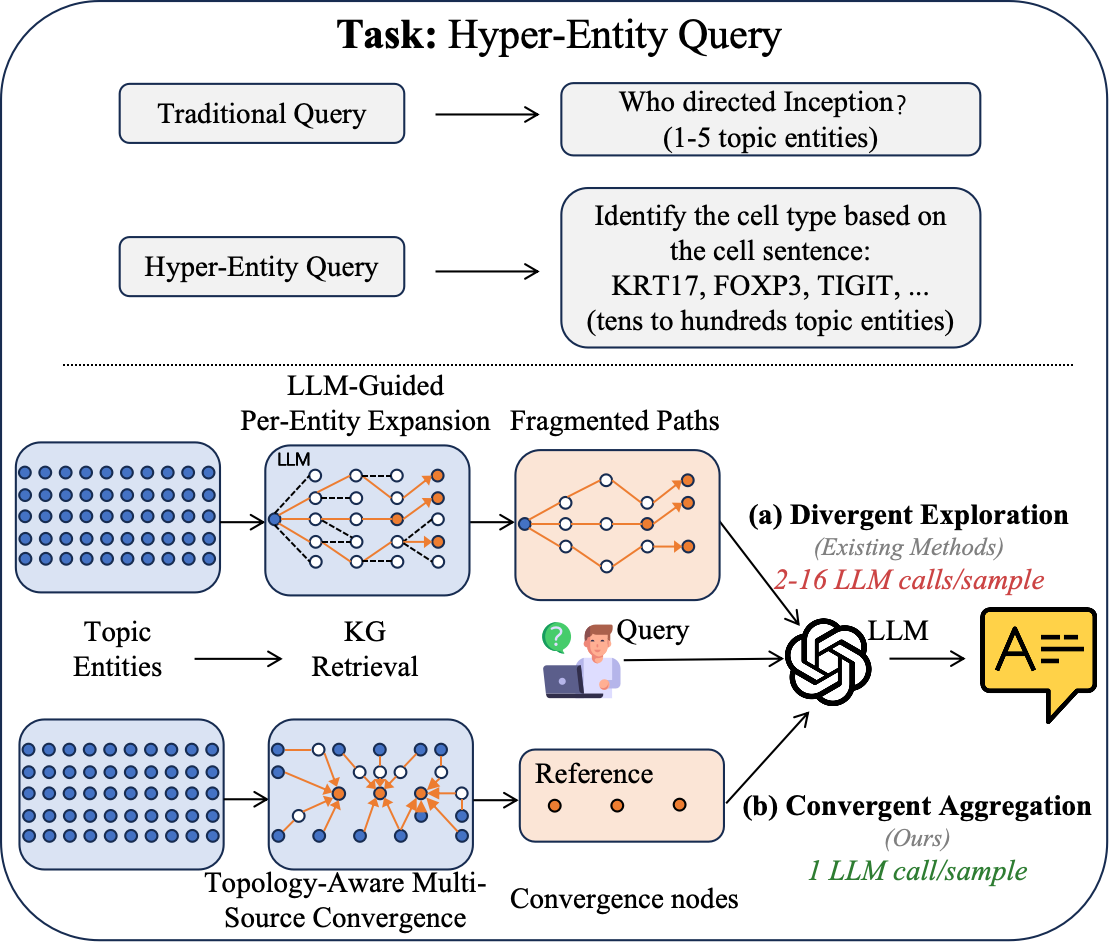}
  \caption{Divergent vs.\ convergent retrieval for hyper-entity queries.
  \textbf{(a)}~Divergent: per-entity LLM-guided expansion.
  \textbf{(b)}~Convergent (GATHER): multi-source traversal identifying topological convergence points.}
  \Description{A comparison between divergent per-entity graph expansion and GATHER's convergent multi-source traversal for hyper-entity queries.}
  \label{fig:framework}
\end{figure}

To address these limitations, we propose \textbf{GATHER} (\textbf{G}raph-\textbf{A}ware \textbf{T}raversal with \textbf{H}yper-\textbf{E}ntity \textbf{R}etrieval), which shifts from local expansion to a global convergence paradigm.
Instead of reasoning from each entity separately, GATHER identifies important \emph{convergence points}---nodes jointly reachable from many source entities---which serve as high-information \emph{hyper-entities}.
These nodes naturally capture the structural interactions among genes and act as consolidated evidence.
Figure~\ref{fig:framework} contrasts this convergent retrieval pattern with divergent per-entity expansion.

We obtain such hyper-entities through a three-stage process.
First, we perform multi-source graph traversal to propagate signals from all input entities simultaneously.
Second, we rank candidate convergence nodes using rank- and topology-aware scores, selecting the most informative hyper-entities.
The selected nodes are then passed to the LLM for final reasoning.

We instantiate GATHER on a self-constructed cell-centric biological knowledge graph and apply it to zero-shot cell-type annotation.
Experiments on two datasets (Immune and Lung) show that GATHER achieves the highest exact-match accuracy (\textbf{27.45\%} and \textbf{59.64\%}) with only one LLM call per sample, outperforming all KG-RAG baselines while using \textbf{2--61$\times$} fewer LLM calls than KG-RAG baselines.
These results demonstrate that convergence nodes compress multi-entity signals into compact, high-information evidence that conveys more per item than multi-hop paths, making convergence-centric retrieval an effective and efficient principle for hyper-entity reasoning.
Code is available at \url{https://github.com/SUAT-AIRI/GATHER}.

\section{Method}
\label{sec:method}

\subsection{Task Formulation}
\label{sec:setup}

We consider zero-shot cell-type annotation under a RAG framework.
We use zero-shot in the training-free sense:
no model is trained or fine-tuned on labeled cells from the evaluation datasets.
The method still relies on curated prior knowledge in VCKG and ranked gene lists,
and is therefore best understood as knowledge-driven, training-free inference.
Given a cell's gene expression profile, we construct a cell sentence
$\mathcal{S}=\{g_1,\ldots,g_n\}$ following Cell2Sentence~\cite{rizvi2025scaling},
where genes are ranked by discriminative power.
We normalize gene symbols against VCKG Gene node symbols and synonyms and map them to canonical Gene nodes.
Uninformative housekeeping genes (e.g., \texttt{RPL*}, \texttt{MT-*}) and symbols without matching nodes are filtered,
yielding the grounded gene set $\tilde{\mathcal{S}}$. The objective is to predict a cell type $c^* \in \mathcal{C}$,
where $\mathcal{C}$ is defined by the Cell Ontology.

In the RAG paradigm, prediction consists of two stages:
(i) retrieving relevant knowledge from a knowledge graph, and
(ii) LLM-based reasoning over the retrieved evidence.

As discussed in the Introduction, hyper-entity queries differ fundamentally from few-entity settings: the correct cell type emerges from the \emph{joint support} of many genes.
Accordingly, we reformulate retrieval as a \emph{multi-source convergence problem}: instead of expanding from each entity independently, the goal is to identify graph nodes jointly supported by multiple genes in $\tilde{\mathcal{S}}$.

\subsection{GATHER: Convergence-Centric Retrieval}
\label{sec:gather}

\begin{figure}[ht]
  \centering
  \includegraphics[width=\columnwidth]{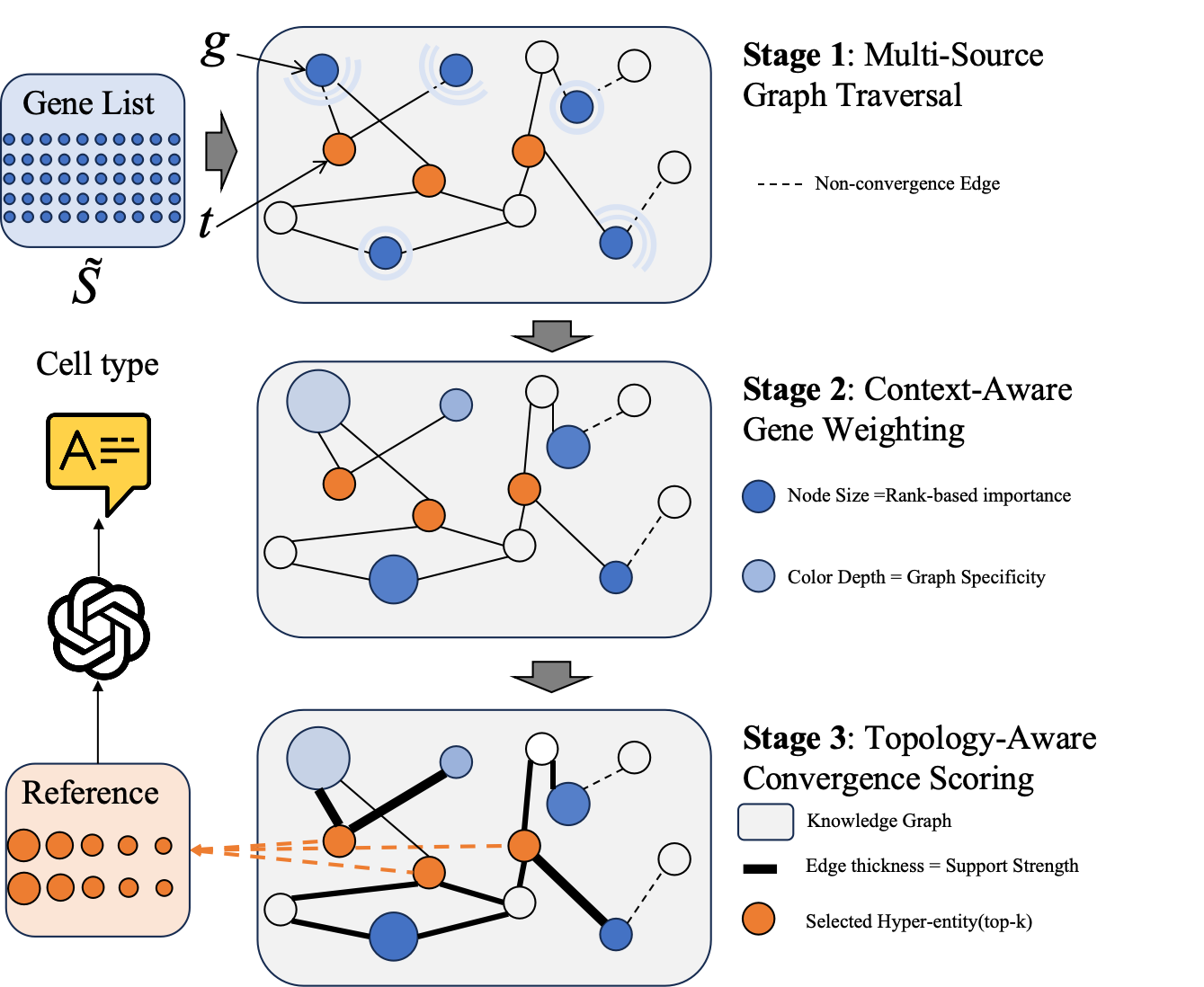}
  \caption{The three stages of GATHER:
  (1)~multi-source graph traversal from the grounded gene set $\tilde{\mathcal{S}}$, where $g$ denotes a source gene and $t$ a candidate target;
  (2)~context-aware gene weighting (node size encodes $w_g^{\text{rank}}$; color depth encodes $w_g^{\text{IDF}}$);
  and (3)~topology-aware convergence scoring, where the top-$K$ candidates ranked by $\mathrm{Score}(t)$ are selected as hyper-entities.}
  \Description{The three algorithmic stages of GATHER: multi-source graph traversal, context-aware gene weighting, and topology-aware convergence scoring.}
  \label{fig:gather-stages}
\end{figure}

Building upon this reformulation, we propose \textbf{GATHER},
a retrieval algorithm designed for hyper-entity queries, as shown in Figure~\ref{fig:gather-stages}.
Rather than performing local expansion from each gene,
GATHER identifies \emph{topological convergence points}---
nodes that receive strong structural support from many source genes.
These nodes act as \emph{hyper-entities}, serving as consolidated evidence
that captures global interactions among genes.
Here, a hyper-entity is not a new biological entity type;
it denotes a retrieved graph node whose relevance is defined by joint support
from a set of source entities rather than by an individual source alone.

GATHER obtains such hyper-entities in three compact stages:
(1)~multi-source traversal, which propagates from grounded genes through the graph and records shared reachability patterns;
(2)~gene weighting, which combines gene rank with graph specificity;
and (3)~convergence scoring, which aggregates hop-binned support to select the final evidence nodes.

\subsubsection{Stage 1: Multi-Source Graph Traversal}

For each $g \in \tilde{\mathcal{S}}$, we traverse the knowledge graph
up to $k$ hops in a relation-agnostic manner (all edge types, both directions).
A semantic-type constraint prevents consecutive nodes of identical type,
avoiding degenerate chains.

Crucially, traversals from all genes proceed simultaneously.
For each discovered target node $t$ (candidate cell-type node),
we record its hop-binned support:

\begin{equation}
S_h(t) = \{ g \in \tilde{\mathcal{S}} \mid g \text{ reaches } t \text{ in exactly } h \text{ hops} \}.
\end{equation}

Nodes jointly reachable from many genes through short paths
naturally emerge as candidate convergence points.

\subsubsection{Stage 2: Context-Aware Gene Weighting}

Stage 1 identifies which genes can support each candidate target,
but raw support counts treat all genes equally.
This is undesirable because top-ranked genes in the cell sentence are more discriminative,
whereas broadly connected genes may reach many targets and provide less specific evidence.
We therefore assign each gene $g$ a combined weight with two components.

\textbf{Rank-based importance:}
\begin{equation}
w_g^{\text{rank}} = \frac{1}{\log_2(\text{rank}(g)+2)}.
\end{equation}

\textbf{Graph specificity (IDF-style):}
\begin{equation}
w_g^{\text{IDF}} =
\log\left(\frac{|\mathcal{T}|}{\text{df}(g)} + 1\right),
\end{equation}

where $\text{df}(g)$ is the number of candidate targets reachable from $g$,
and $\mathcal{T}$ is the union of all discovered targets.

The rank term favors salient genes in the cell sentence,
whereas the IDF term downweights genes that reach many candidate targets.
Together, they make the subsequent convergence score depend on selective,
high-rank support rather than raw reachability.

\subsubsection{Stage 3: Topology-Aware Convergence Scoring}

We rank each candidate node $t$ by aggregating weighted support:

\begin{equation}
\text{Score}(t)
=
\sum_{h=1}^{k}
\alpha_h
\sum_{g \in S_h(t)}
w_g^{\text{rank}} \cdot w_g^{\text{IDF}},
\end{equation}

where $\alpha_h$ are hop-decay weights favoring shorter paths.

This scoring function directly operationalizes the convergence principle:
nodes jointly supported by many informative and specific genes
through short paths receive higher scores.

The top-$K$ convergence nodes,
together with their supporting gene coalitions,
form a compressed evidence context
that is passed to a single LLM call for final reasoning.

Compared to entity-wise local expansion,
GATHER provides:
(1) \textbf{Global coordination:}
    explicitly modeling joint structural support across genes;
(2) \textbf{Context compression:}
    distilling $N$ source genes into $K \ll N$ convergence nodes;
(3) \textbf{LLM cost reduction:}
    requiring zero LLM calls during retrieval.
Thus, retrieval shifts from fragmented local exploration
to structured global aggregation.

The retrieval cost is governed by the number of grounded genes,
the traversal horizon, and the local graph fan-out.
In the worst case, a naive traversal grows as $O(|\tilde{\mathcal{S}}|d^k)$
for average fan-out $d$, but GATHER uses a shallow fixed horizon
and ranks only candidate cell-type targets.
The IDF term is therefore traversal-specific:
$\text{df}(g)$ is computed under the traversal horizon $k$ used for a given run,
rather than as a global gene frequency or a per-sample tuned quantity.

\subsection{VCKG as a Cell-Centric Knowledge Graph for Hyper-Entity Retrieval}
\label{sec:vckg}

GATHER requires a knowledge graph where genes serve as entry points,
functional and cellular semantics enable multi-hop paths to cell types,
and cell-type nodes are grounded in a formal ontology.
No existing public KG satisfies all three requirements:
general-purpose biomedical KGs (e.g., PrimeKG~\cite{chandak2023building}) lack a cell-centric schema,
while domain-specific resources (e.g., CellMarker~\cite{zhang2019cellmarker})
are flat databases without the graph topology needed for multi-hop convergence.

We therefore construct \textbf{VCKG},
a cell-centric biological knowledge graph
that integrates 20+ databases and 7 domain ontologies
through a four-step pipeline:
(1)~\emph{data collection} from sources spanning genes (NCBI Gene, HGNC, UniProt),
functions (GO~\cite{ashburner2000gene}, Reactome),
cells (Cell Ontology, CellMarker 2.0~\cite{hu2023cellmarker}),
anatomy (UBERON), and diseases (DO, MONDO, HPO);
(2)~\emph{ontology normalization}, mapping every entity to its canonical identifier
to resolve synonyms and cross-database conflicts;
(3)~\emph{relation standardization} via the Relation Ontology;
and (4)~\emph{graph assembly} into a Neo4j property graph.
To reduce cross-source ambiguity, gene symbols and aliases are normalized
against HGNC and NCBI Gene identifiers, protein-level references are linked
through UniProt accessions, and cell types are grounded to Cell Ontology IDs.
When multiple sources describe the same entity, we merge by canonical identifier
and preserve source-specific fields, such as \texttt{field\_sources},
\texttt{source}, or PMID metadata, when available.
Because direct marker annotations are incomplete, VCKG supports both direct
\texttt{IS\_MARKER\_FOR} evidence and indirect functional or ontological paths,
allowing GATHER to exploit convergence beyond one-hop marker lookup.

\begin{table}[ht]
\centering
\caption{Key statistics of VCKG.}
\label{tab:vckg_stats}
\begin{tabular}{lr|lr}
\toprule
\textbf{Metric} & \textbf{Value} & \textbf{Metric} & \textbf{Value} \\
\midrule
Total Nodes & 120K+ & Total Edges & 2,500K+ \\
Node Types & 9 & Edge Types & 14 \\
Ontologies & 7 & Data Sources & 20+ \\
Avg. Node Degree & $\sim$21 & Max Path Length & 5 \\
\bottomrule
\end{tabular}
\end{table}

The key design principle is \emph{gene-centric structural connectivity}:
43,000+ gene nodes act as hubs linking to functional concepts (GO terms, pathways)
and 2,500+ Cell Ontology cell-type nodes.
A gene connects to a cell type either directly
via \texttt{IS\_MARKER\_FOR} (1-hop)
or indirectly through shared annotations,
e.g., $\texttt{Gene} \xrightarrow{\texttt{PARTICIPATES\_IN}} \texttt{BP} \xleftarrow{\texttt{CAPABLE\_OF}} \texttt{CellType}$ (2-hop),
enabling convergence signals from both direct marker evidence
and indirect functional overlap.
Table~\ref{tab:vckg_stats} summarizes the key statistics.

\section{Experiments}

\subsection{Experimental Setup}

\paragraph{Datasets.}
We evaluate on two single-cell datasets whose labels adopt standardized Cell Ontology nomenclature, enabling direct alignment with VCKG's ontology-grounded cell type nodes:
(1)~\textbf{Immune Human}~\cite{dominguez2022cross}: a subset from the cross-tissue immune cell atlas (donors A29 and A31, following the Cell2Sentence~\cite{rizvi2025scaling} split), comprising 2,962 test cells across 34 fine-grained types;
(2)~\textbf{Tabula Sapiens Lung}~\cite{the2022tabula}: the lung tissue partition from the Tabula Sapiens multi-organ atlas, comprising 2,416 test cells across 30 types.
Each cell is converted to a cell sentence of $n{=}50$ ranked genes (source entities), and the number of retained convergence nodes is set to $K{=}10$.
In all main experiments, we set the traversal horizon to $k{=}2$
to balance coverage and specificity,
while the GATHER formulation supports other traversal horizons.

\paragraph{Baselines.}
We compare GATHER against representative baselines spanning direct LLM prompting and diverse KG-based RAG paradigms.
\textbf{LLM} performs direct prompting on gene lists without external knowledge.
Among KG-based methods, existing approaches predominantly adopt a \emph{local, entity-wise expansion} paradigm, where each gene is explored independently and evidence is aggregated afterward.
\textbf{RoG}~\cite{luo2024rog} follows a template-based strategy by predicting relation paths and performing per-entity traversal.
\textbf{ToG}~\cite{sun2024tog} and its enhanced variant \textbf{ToG-2}~\cite{ma2025tog2} perform iterative, LLM-guided entity-level exploration with dynamic pruning.
\textbf{PoG}~\cite{tan2025paths} conducts multi-stage path-centric search with LLM-based refinement.
All methods use the same LLM backbone (GPT-4o-mini) and VCKG as the knowledge source for fair comparison.

\paragraph{Metrics.}
We report exact-match accuracy, average LLM calls per sample, and evidence quantity (average number of evidence items retrieved per sample).
To account for the hierarchical nature of cell types, we additionally evaluate against the Cell Ontology DAG:
Ancestor Match~(Anc.) credits predictions that lie on the same root-to-leaf path as the true label (i.e., one is an ancestor or descendant of the other). We focus on exact and ontology-aware matching metrics, as they directly reflect whether retrieval succeeds in identifying the correct semantic region in the Cell Ontology, rather than smoothing errors through label averaging.

\begin{table}[ht]
\centering
\small
\caption{Main results on cell-type annotation.
Exact: exact-match accuracy (\%);
Anc.: ancestor match (\%), where a prediction is credited if it is an ancestor or descendant of the true label in the Cell Ontology;
Calls: average LLM calls per sample;
Evid.: average evidence items per sample (convergence nodes for GATHER; relation paths for RoG; reasoning triples for ToG/ToG-2; retrieved paths for PoG).
All methods use GPT-4o-mini and VCKG.}
\label{tab:main_results}
\resizebox{\linewidth}{!}{
\begin{tabular}{lcccccccc}
\toprule
& \multicolumn{4}{c}{\textbf{Immune} (34 types)} 
& \multicolumn{4}{c}{\textbf{Lung} (30 types)} \\
\cmidrule(lr){2-5} \cmidrule(lr){6-9}
\textbf{Method} 
& \textbf{Exact} & \textbf{Anc.} & \textbf{Calls} & \textbf{Evid.}
& \textbf{Exact} & \textbf{Anc.} & \textbf{Calls} & \textbf{Evid.} \\
\midrule
LLM & 14.01 & 26.40 & 1.0 & --- & 54.88 & 54.97 & 1.0 & --- \\
RoG & 17.39 & 31.53 & 2.0 & 12.8 & 56.37 & 56.79 & 2.0 & 24.8 \\
ToG-2 & 18.13 & 29.03 & 13.3 & 18.8 & 53.35 & 53.39 & 12.6 & 10.9 \\
PoG & 20.80 & 33.15 & 15.9 & 6.8 & 48.59 & 48.80 & 8.2 & 7.8 \\
ToG & 20.50 & 36.61 & 56.2 & 20.0 & 56.04 & 56.21 & 60.5 & 20.0 \\
\midrule
\textbf{GATHER} 
& \textbf{27.45} & 33.09 & \textbf{1.0} & 9.3
& \textbf{59.64} & \textbf{60.18} & \textbf{1.0} & 10.0 \\
+Path 
& 26.54 & 31.94 & 1.0 & 9.3
& 58.65 & 59.19 & 1.0 & 10.0 \\
\bottomrule
\end{tabular}
}
\end{table}

\subsection{Main Results}

\paragraph{Overall Performance and Efficiency.}
From Table~\ref{tab:main_results}, GATHER achie-ves the highest exact-match accuracy on both datasets:
\textbf{27.45\%} on Immune (vs.\ 20.80\% for PoG) and
\textbf{59.64\%} on Lung (vs.\ 56.37\% for RoG).
Notably, these gains are obtained with only 1 LLM call per sample,
whereas KG-RAG baselines require between 2.0 and 60.5 LLM calls.
Although increased LLM interaction generally correlates with improved accuracy
(e.g., ToG-2: 13.3 calls $\to$ 18.13\% on Immune; ToG: 56.2 calls $\to$ 20.50\%),
GATHER surpasses all baselines without additional LLM reasoning steps.
Beyond LLM efficiency, GATHER also achieves \emph{evidence compression}.
Although GATHER retrieves 9.3--10.0 evidence items,
comparable in count to baselines such as PoG (6.8--7.8 paths),
the information granularity differs fundamentally:
each baseline evidence item is a \emph{multi-hop path} consisting of several nodes and edges
(e.g., a 2-hop path contains 3 nodes and 2 relations),
whereas each GATHER evidence item is a single \emph{convergence node}
that aggregates structural support from multiple source genes.
Thus, in terms of total information volume fed to the LLM,
GATHER uses substantially less knowledge than path-based methods
while achieving the highest exact-match accuracy.
This demonstrates that convergence modeling produces
inherently more informative evidence per item,
and that retrieval quality---not quantity---drives performance.

\paragraph{Cross-Dataset Analysis.}
Performance patterns differ between the two datasets.
Lung yields higher absolute accuracy for all methods
(e.g., LLM baseline: 54.88\% on Lung vs.\ 14.01\% on Immune),
reflecting its relatively coarser and more transcriptionally distinct cell types.
Immune, containing fine-grained T-cell subtypes,
is more sensitive to multi-gene interactions.
Despite this increased difficulty,
GATHER maintains the best exact-match accuracy in both regimes.
On Lung, it also achieves the highest ancestor match (60.18\%),
indicating correct placement within the ontology hierarchy.
On Immune, while the ancestor match (33.09\%) is below ToG (36.61\%), the higher exact accuracy suggests that GATHER favors precise subtype predictions over conservative coarse-grained guesses.
These observations support that convergence modeling is particularly beneficial in fine-grained hyper-entity settings, where the label cannot be resolved from any single gene and must instead emerge from multi-gene structural agreement.

\paragraph{Ablation and Scaling Analysis.}
The path ablation further clarifies the effective signal.
When explicit traversal paths are added,
accuracy decreases from 27.45\% to 26.54\% on Immune
and from 59.64\% to 58.65\% on Lung.
This suggests that the LLM primarily benefits from
\emph{which genes converge on a candidate and at what distance},
rather than verbose path descriptions,
reinforcing that topological convergence is the core signal.

Gene-scaling results (Table~\ref{tab:gene_scaling})
show that exact-match accuracy generally improves
as the number of input genes grows.
On Lung, accuracy rises from 55.46\% (10 genes)
to 59.64\% (50 genes);
on Immune, it improves from 24.85\% (10 genes)
to 27.72\% (40 genes) and remains comparable at 27.45\% with 50 genes.
This trend suggests that additional genes strengthen the convergence signal until it saturates rather than dilute it.
In contrast, traditional entity-wise methods
do not exhibit comparable scaling gains,
as their retrieval treats each gene independently.
Finally, varying the number of retained convergence nodes
($K \in \{5,10,15\}$) yields similar performance
(Immune: 26.87\%/27.45\%/27.04\%; Lung: 59.48\%/59.64\%/59.81\%),
with $K{=}10$ achieving a good balance across both datasets,
indicating that convergence signals are stably concentrated
among a small set of top candidates.

\begin{table}[h]
\centering
\small
\caption{Effect of input gene count on GATHER (GPT-4o-mini). Genes: number of input genes from the cell sentence; Grounded: genes successfully mapped to VCKG nodes; Exact: exact-match accuracy (\%).}
\label{tab:gene_scaling}
\begin{tabular}{ccccc}
\toprule
& \multicolumn{2}{c}{\textbf{Immune}} & \multicolumn{2}{c}{\textbf{Lung}} \\
\cmidrule(lr){2-3} \cmidrule(lr){4-5}
\textbf{Genes} & \textbf{Grounded} & \textbf{Exact} & \textbf{Grounded} & \textbf{Exact} \\
\midrule
10 & 10.0 & 24.85 & 10.0 & 55.46 \\
20 & 20.0 & 26.03 & 20.0 & 57.12 \\
30 & 30.0 & 26.84 & 30.0 & 58.11 \\
40 & 40.0 & 27.72 & 40.0 & 59.44 \\
50 & 48.8 & 27.45 & 47.7 & 59.64 \\
\bottomrule
\end{tabular}
\end{table}
\balance
\subsection{Discussion and Limitations}

\paragraph{Scope of Comparison.}
This work focuses on training-free KG-RAG retrieval rather than supervised
cell-type classification.
Non-KG methods such as linear classifiers, PCA-based workflows,
set-based encoders, and graph neural networks
are important complementary baselines,
but they typically require labeled cells, feature learning,
or task-specific training.
Comparing these paradigms under a unified data and supervision protocol
is an important direction for future work.

\paragraph{Dependence on VCKG.}
GATHER assumes that the underlying graph contains meaningful convergence targets
and that cell-type nodes are sufficiently grounded in ontology and marker evidence.
Its performance may degrade when the true cell type is absent from the graph,
when marker coverage is sparse, or when input genes cannot be reliably grounded.
Applying the method to a general biomedical KG such as PrimeKG would require
additional schema alignment, target typing, and cell-type grounding,
because such KGs are not organized around cell-type retrieval.

\paragraph{Evidence and Failure Modes.}
Convergence nodes should be interpreted as retrieval evidence,
not as direct proof of a biological mechanism.
On fine-grained immune subtypes, GATHER improves exact match, but can still
make overly specific errors, as reflected by its lower ancestor match than ToG.
Future work should report top-$K$ retrieval ceilings, per-class confusion patterns,
expert validation of retrieved convergence nodes,
and robustness to missing genes or noisy graph edges.

\paragraph{Parameter Sensitivity.}
Our main experiments use $n{=}50$ input genes
and a two-hop traversal horizon as a compact default setting.
The gene-count analysis suggests that convergence signals strengthen
once sufficient source genes are available, but the best traversal depth
and relation constraints may vary across knowledge graphs and annotation
granularity.
Future work should study when additional graph context improves coverage
or introduces overly broad or noisy convergence targets.

\section{Conclusion}
We introduced \textbf{GATHER}, a convergence-centric retrieval framework
for hyper-entity queries in KG-RAG,
where answers emerge from the collective support of many entities.
Instead of independent entity-wise expansion,
GATHER detects topological convergence points
to model multi-source structural synergy during retrieval.
Across two benchmarks, GATHER achieves the best exact-match accuracy
(27.45\% on Immune and 59.64\% on Lung)
while requiring only a single LLM call per sample,
reducing LLM usage by 2--61$\times$ compared to KG-RAG baselines.
These results demonstrate that convergence nodes compress collective entity signals
into compact, high-information evidence,
and that improving retrieval quality,
rather than increasing evidence volume or iterative LLM reasoning,
is key to effective hyper-entity modeling.
More broadly, our findings highlight evidence informativeness as a key factor
in zero-shot biomedical reasoning with collective entity signals.


\begin{acks}
This work was supported by the project of Shenzhen Application Research and Development Special Fund Support (Grant No. XLQSQ20250427092505008), the National Key Research and Development Program of China (Grant No. 2024YFF0908200), the Natural Science Foundation of Guangdong Province of China (Grant Nos. 2024A1515030166 and 2025B1515020032), the Shenzhen Science and Technology Innovation Program (Grant No. KQTD20190929172835-662), and the Innovation Team Project of Guangdong Province (Grant No. 2024KCXTD017).
\end{acks}


\bibliographystyle{ACM-Reference-Format}

\bibliography{references}

\end{document}